\documentclass[10pt,twocolumn,letterpaper]{article}

\usepackage{cvm}
\usepackage{times}
\usepackage{epsfig}
\usepackage{graphicx}
\usepackage{amsmath}
\usepackage{amssymb}
\usepackage{authblk}

\usepackage{bm}
\usepackage[pagebackref=true,breaklinks=true,letterpaper=true,colorlinks,bookmarks=false]{hyperref}

\cvmfinalcopy
\def\cvmPaperID{****} % *** Enter the cvm Paper ID here

\ifcvmfinal\fi
\begin{document}

%%%%%%%%% TITLE
\title{Convolutional Prototype Learning for Zero-Shot Recognition}
\author[1,2]{Zhizhe Liu}
\author[1,2]{Xingxing Zhang}
\author[1,2]{Zhenfeng Zhu}
\author[1,2]{ShuaiZheng}
\author[1,2]{YaoZhao}
\author[3]{JianCheng}
% \affil[1]{Institute of Information Science, Beijing Jiaotong University, Beijing, China}
% \affil[2]{Beijing Key Laboratory of Advanced Information Science and Network Technology, Beijing, China}
% \affil[3]{NLPR, Institute of Automation, Chinese Academy of Sciences, Beijing, China }
% \renewcommand\Authands{ and }
\affil[1]{Institute of Information Science, Beijing Jiaotong University, Beijing, China}
\affil[2]{Beijing Key Laboratory of Advanced Information Science and Network Technology, Beijing, China}
\affil[3]{NLPR, Institute of Automation, Chinese Academy of Sciences, Beijing, China  \authorcr \texttt { \{zhz1iu, zhangxing, zhfzhu, zs1997, yzhao\}@bjtu.edu.cn, jcheng@nlpr.ia.ac.cn}}

\renewcommand\Authands{ and }

% \texttt \authorcr

% \affil[*]{单位1, \authorcr Email: \{zuozhe1, zuozhe2\}@yahoo.com, zuozhe3@sina.com}

%{\tt\small {zhzliu,zhangxing,zhfzhu,zs1997,yzhao}@bjtu.edu.cn, jcheng@nlpr.ia.ac.cn}

\maketitle
% \thispagestyle{empty}

%%%%%%%%% ABSTRACT
\begin{abstract}
    Zero-shot learning (ZSL) has received increasing attention in recent years especially in areas of fine-grained object recognition, retrieval, and image captioning. The key to ZSL is to transfer knowledge from the seen to the unseen classes via auxiliary class attribute vectors. However, the popularly learned projection functions in previous works cannot generalize well since they assume the distribution consistency between seen and unseen domains at sample-level. 
    Besides, the provided non-visual and unique class attributes can significantly degrade the recognition performance in semantic space. In this paper, we propose a simple yet effective convolutional prototype learning (CPL) framework for zero-shot recognition. By assuming distribution consistency at task-level, our CPL is capable of transferring knowledge smoothly to recognize unseen samples.
    Furthermore, inside each task, discriminative visual prototypes are learned via a distance based training mechanism. Consequently, we can perform recognition in visual space, instead of semantic space. An extensive group of experiments are then carefully designed and presented, demonstrating that CPL obtains more favorable effectiveness, over currently available alternatives under various settings.
       
\end{abstract}

%%%%%%%%% BODY TEXT
%------------------------------Introduction---------------------------------------------------------%
\section{Introduction}
%%
%  In numerous practical applications, we need the model to have the ability to determine the class labels for the data belonging to unseen classes. The following are some popular application scenarios~\cite{wang2019survey}:

 %----------------Figure 1------------------------------------%
 \begin{figure}[t!]
 % \vskip -0.2in
 \begin{center}
 \centerline{\includegraphics[width=0.5\textwidth,height=0.28\textwidth]{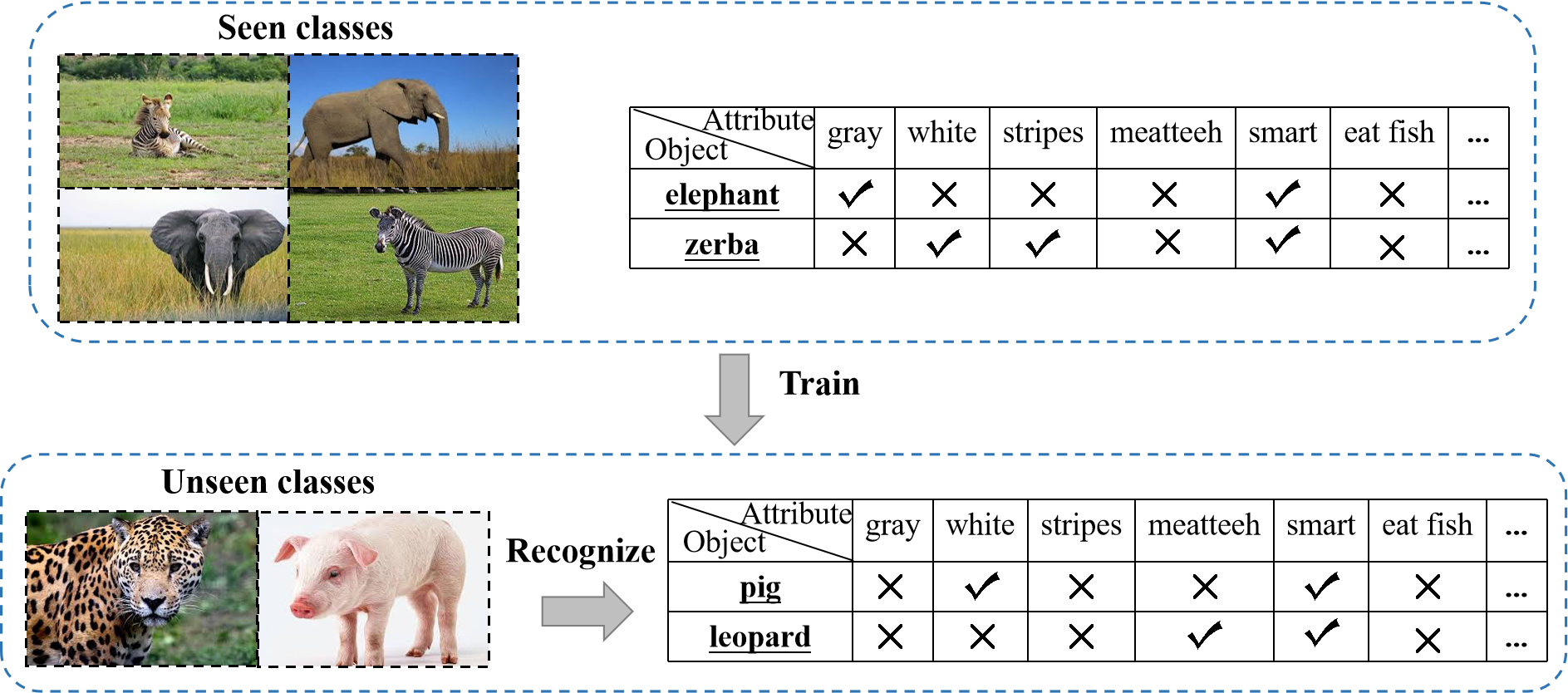}}
 \caption{The visual images and class prototypes provided for several classes in benchmark dataset AWA2~\cite{xian2018zero}.}
 \label{fig:att_image}
 \end{center}
 \vskip -0.4in
 \end{figure}
 %----------------Figure 1------------------------------------%
 
In recent years, deep convolutional neural networks have achieved significant pro-gress in object classification.
However, most existing recognition models need to collect and annotate a large amount of target class samples for model training.
It is obvious these operations are expensive and cumbersome.
In addition, the number of target classes is very large, and meanwhile novel categories appear dynamically in nature daily.
Moreover, for fine-grained object recognition (e.g., birds), it is hard to collect enough image samples for each category (e.g., skimmer) due to the rarity of target class.
Thus, zero-shot learning (ZSL)~\cite{mensink2014costa,zhang2015zero, changpinyo2016synthesized,morgado2017semantically,xian2018feature,felix2018multi,schonfeld2019generalized,kampffmeyer2019rethinking} is proposed to address above issues.

ZSL aims to recognize objects which may not have seen instances in the training phase.
Due to the lower requirement of labeled samples and various application scenarios, it recently has received a lot of focus and achieved significant advances in computer vision.
Because of the lack of labeled samples in the unseen class domain, ZSL requires the auxiliary information to learn the transferable knowledge from the seen to unseen class domain.
To this end, existing methods usually offer the semantic description extracted from text (e.g., attribute vector~\cite{kankuekul2012online, lampert2009learning, lampert2014attribute} or word vector~\cite{akata2015evaluation, frome2013devise, socher2013zero}) for each category to relate the seen and unseen class domains as shown in Figure~\ref{fig:att_image}.
This mimics the human ability to recognize the novel objects in the world.
For example, given the description that ``a wolf looks like a dog, but has a drooping tail'', we can recognize a wolf even without having seen one, as long as we have seen a ``dog".
It is obvious zero-shot learning is feasible by imitating such a learning mechanism.

  %--------------------Figure 2---------------------------%
  \begin{figure}[t!]
  % \vskip -0.2in
  \begin{center}
  \centerline{\includegraphics[width=2.9in]{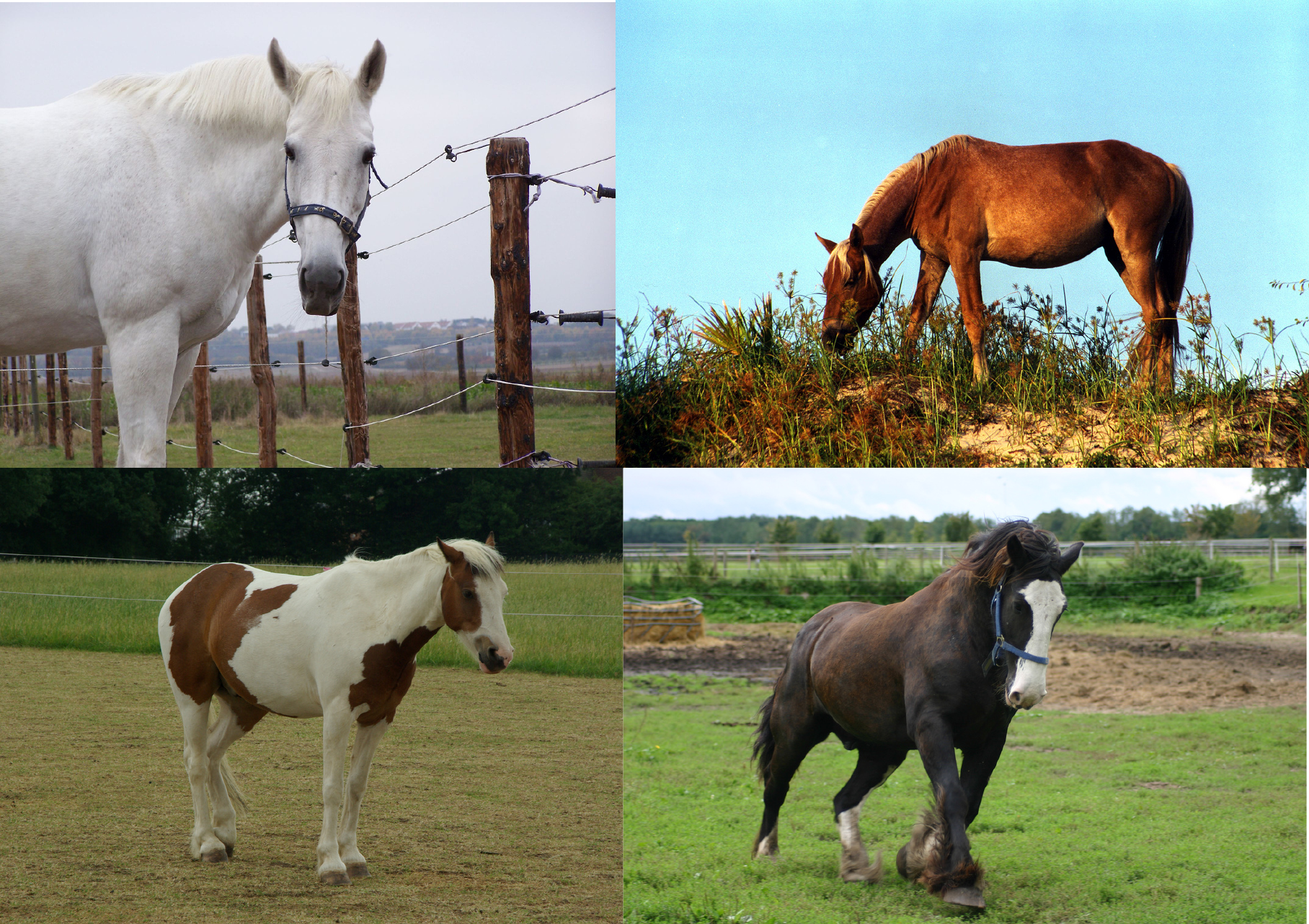}}
  \caption{Several instances with various colors from the unseen class ``horse'' in benchmark dataset AWA2~\cite{xian2018zero}.}
  \label{fig:horse}
  \end{center}
  \vskip -0.3in
  \end{figure}
  %--------------------Figure 2---------------------------%
 
 The key of ZSL is how effectively learn the visual-semantic projection function.
 Once the visual-semantic projection function is learned from the source object domain, we can transfer the learned knowledge to target object domain.
 At test phase, the test sample is first projected into the embedding space, and then the recognition is conducted by computing the similarity between the test sample and unseen class attributes.
 To this end, various ZSL methods are proposed to learn an effective projection function between visual and semantic spaces recently.
 However, due to there exists a significant distribution gap between the seen and unseen samples in most real scenarios, the generalization ability of such methods is seriously limited.
 For example, both ``zerba'' and ``pig'' have the attribute ``tail'', but the visual appearance of their tails is often greatly different.
 To solve this problem, a lot of ZSL methods propose transductive learning~\cite{fu2015transductive,guo2016transductive,niu2018webly,zhao2018domain} which introduces visual samples of unseen classes in the training phase.
 Obviously, transductive learning can mitigate the domain shift, however it is hard to apply into many real scenarios due to the great challenge to obtain the corresponding samples for all target classes.
 In this paper, we aim at developing an inductive ZSL method that can reduce the domain distribution mismatch between the seen and the unseen class data, only by training set.
%--------------------Figure 3---------------------------%
 \begin{figure}[t!]
 \vskip -0.15in
 \begin{center}
 \centerline{\includegraphics[width=\columnwidth]{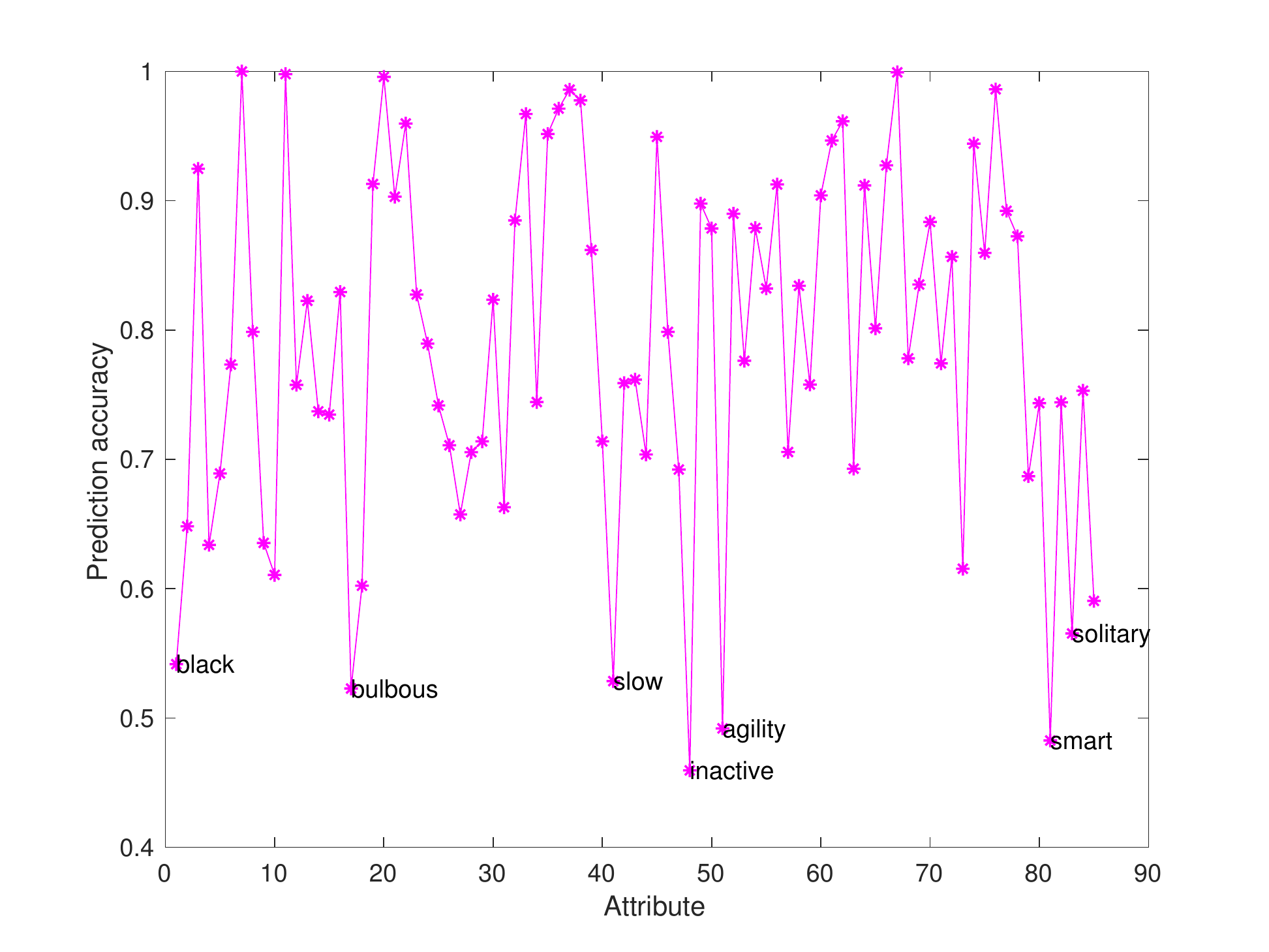}}
 \caption{The predictability of each binary attribute, measured with classification accuracy by pre-trained Resnet101~\cite{he2016deep}, offered by~\cite{1910.11671}.}
 \label{error}
 \end{center}
 \vskip -0.3in
 \end{figure}
 %--------------------Figure 3---------------------------%

In addition, both handly defined attribute vectors and automatically extracted word vectors in ZSL are obtained independently from visual samples, and uniquely for each class.
This results in the class descriptors are usually inaccurate and less diversified.
Due to this reason, the projection learned in previous works may not effectively address the intra-class variation problem.
For instance, there are many colors for a ``horse'' as shown in Figure~\ref{fig:horse}.
Besides, there exists some non-visual attributes in the provided class descriptors, such as ``timid'', ``solitary'', and ``active'' in benchmark dataset AWA2~\cite{xian2018zero}.
Obviously, these attributes are hard to be predicted only based on visual information.
Referring to~\cite{1910.11671}, it even may appear a level of random guess as shown in Figure~\ref{error}.
Thus, the learned knowledge cannot effectively transfer to the unseen class domain, although we can achieve significant performance in the seen class domain thanks to supervised learning.
Specially, due to different shooting angles, the visual instance of an object generally cannot possess all the attributes provided by class descriptor.
As a result, a common projection from visual to semantic space usual is inaccurate since the lack of some attributes (e.g., paws and tail are not captured).

 Motivated by these two key observations above, in this work, we focus on developing a novel inductive ZSL model. Different from previous works, we assume distribution consistency between the seen and unseen class samples at task-level, instead of sample-level, for better transferring knowledge.
 Furthermore, considering the non-visual components and uniqueness of provided class descriptors, we choose to learn discriminative visual prototypes, thus performing zero-shot recognition in visual space, instead of semantic space.

 We emphasize our \textbf{contributions} in four aspects:
 \begin{itemize}
 \item[-] A simple yet effective convolutional prototype learning (CPL) framework is proposed for zero-shot recognition task.
 \item[-] Our CPL is able to transfer knowledge smoothly with an assumption of distribution consistency at task-level, thus recognizing unseen samples more accurately.
 \item[-] To avoid the problems of recognition in semantic space, discriminative visual prototypes are learned via a distance based training mechanism in our CPL.
 \item[-] The improvements over currently available ZSL alternatives are especially significant under various ZSL settings.
 \end{itemize}

 \section{Related work}
 According to whether information about the test data is involved during model learning, existing ZSL models consist of inductive~\cite{annadani2018preserving, changpinyo2016synthesized, kodirov2017semantic, romera2015embarrassingly} and transductive settings~\cite{fu2015transductive,fu2018zero,kodirov2015unsupervised,niu2018webly}. Specifically, this transduction in ZSL can be embodied in two progressive degrees: transductive for specific unseen classes~\cite{liu2018generalized} and transductive for specific test samples~\cite{zhao2018domain}. Although transductive settings can alleviate the domain shift caused by the different distributions of the training and the test samples, it is not realistic to obtain all test samples. Thus, we take an inductive setting in this work.
 
 From the view of constructing the visual-semantic interactions, existing inductive ZSL methods fall into four categories. 
 The first group learns a projection function from the visual to the semantic space with a linear~\cite{bansal2018zero, lampert2014attribute, li2018discriminative} or a non-linear model~\cite{chen2018zero,  morgado2017semantically, socher2013zero, yu2018stacked}. The test unseen data are then classified by matching the visual representations in the class semantic embedding space with the unseen class prototypes. To capture more distribution information from visual space, recent work focuses on generating pseudo examples for unseen classes with seen class examples~\cite{guo2017zero}, web data~\cite{niu2018webly}, generative adversarial networks (GAN)~\cite{xian2018feature,zhu2018generative}, etc. Then supervised learning methods can be employed to perform recognition task.
 The second group chooses the reverse projection direction~\cite{annadani2018preserving,wang2018zero,zhang2017learning}, to alleviate the hubness problem caused by nearest neighbour search in a high dimensional space~\cite{radovanovic2010hubs}. The test unseen data are then classified by resorting to the most similar pseudo visual examples in the visual space. 
 The third group is a combination of the first two groups by taking the encoder-decoder paradigm but with the visual feature or class prototype reconstruction constraint~\cite{annadani2018preserving, kodirov2017semantic}. It has been verified that the projection function learned in this way is able to generalize better to the unseen classes. 
 The last group mainly learns an intermediate space, where both the visual and the semantic spaces are projected to~\cite{changpinyo2018classifier, changpinyo2017predicting, hubert2017learning,liu2018generalized}. 
 
 To avoid the problems caused by non-visual components and uniqueness of the provided class attributes in ZSL, we choose to perform recognition in visual space. Thus, our CPL can be considered as one of the second group.
 
%-------------------------------------Method--------------------------------------------%
\section{Methodology}
In this section, we first set up the zero-shot recognition~(Section~\ref{Setup}), then develop a convolutional prototype learning (CPL) framework for this task~(Section~\ref{ModelFormulation}), and finally discuss how to perform recognition on test data~(Section~\ref{ModelPrediction}).

\subsection{Problem definition} \label{Setup}

 Suppose we have an unseen class set $Y^u =\left \{u_1,\cdots,u_L\right \}$, where there are not any labeled samples for these $L$ different classes.
 But each class is provided with an attribute vector correspondingly, and then we denote $\bm A^{u}=\left [ \bm a_{1}^{u},\cdots,\bm a_{L}^{u}  \right ]$ as all unseen class attribute vectors.
 Given an unlabeled sample set $\mathcal{X}^{te}$, the target of zero-shot learning is to infer the correct class label from the $L$ classes for each sample in $\mathcal{X}^{te}$.
 It is impossible to learn an effective classifier only from these data. 
 To tackle this problem, an additional training set $\mathcal{D}^{tr}=\left \langle\mathcal{X}^{tr}, \mathcal{Y}^{tr}\right \rangle$ that covers $K$ seen classes, is usually adopted to learn transferable knowledge to help the classification on $\mathcal{X}^{te}$. 
 $\mathcal{X}^{tr}$ and $\mathcal{Y}^{tr}$ represent the sets of training samples and corresponding labels, respectively.
 Let $Y^s=\left \{ s_1,\cdots,s_K \right \} $ denote the set of these $K$ seen classes, and $ \bm A^{s}=\left [ \bm a_{1}^{s},\cdots,\bm a_{K}^{s}\right]$ is corresponding attribute vectors.
 It is worth noting that, generally, $K>L$. More importantly, $Y^u \cap Y^s =\emptyset$ under the standard ZSL setting, while $Y^u\subseteq Y^s$ under the generalized ZSL setting.
 
 Let $\Psi (x)$ denote a convolutional neural network based embedding module, which can learn feature representation for any input image $x$. We use $f(\Psi (x), \bm A^{u}): \mathbb{R}^{d}\to\mathbb{R}^{L}$ as a classifier moduler to assign a label $y$ for a test image $x$ within $L$ different classes, according to the unseen class attribute set $\bm A^{u}$. Denote the true class label of $x$ as $y_{true}$. Note that these two modules can be integrated into a unified network and trained from scratch in an end-to-end manner.
 The cost function of $f(\Psi (x), \bm A^{u})$ is denoted as $\mathcal{L}(x, \bm A^{u}, y_{true})$ for simplicity.

\subsection{CPL: algorithm}\label{ModelFormulation}
\begin{figure}[t!]
 \vskip -0.15in
 \begin{center}
 \centerline{\includegraphics[width=\columnwidth]{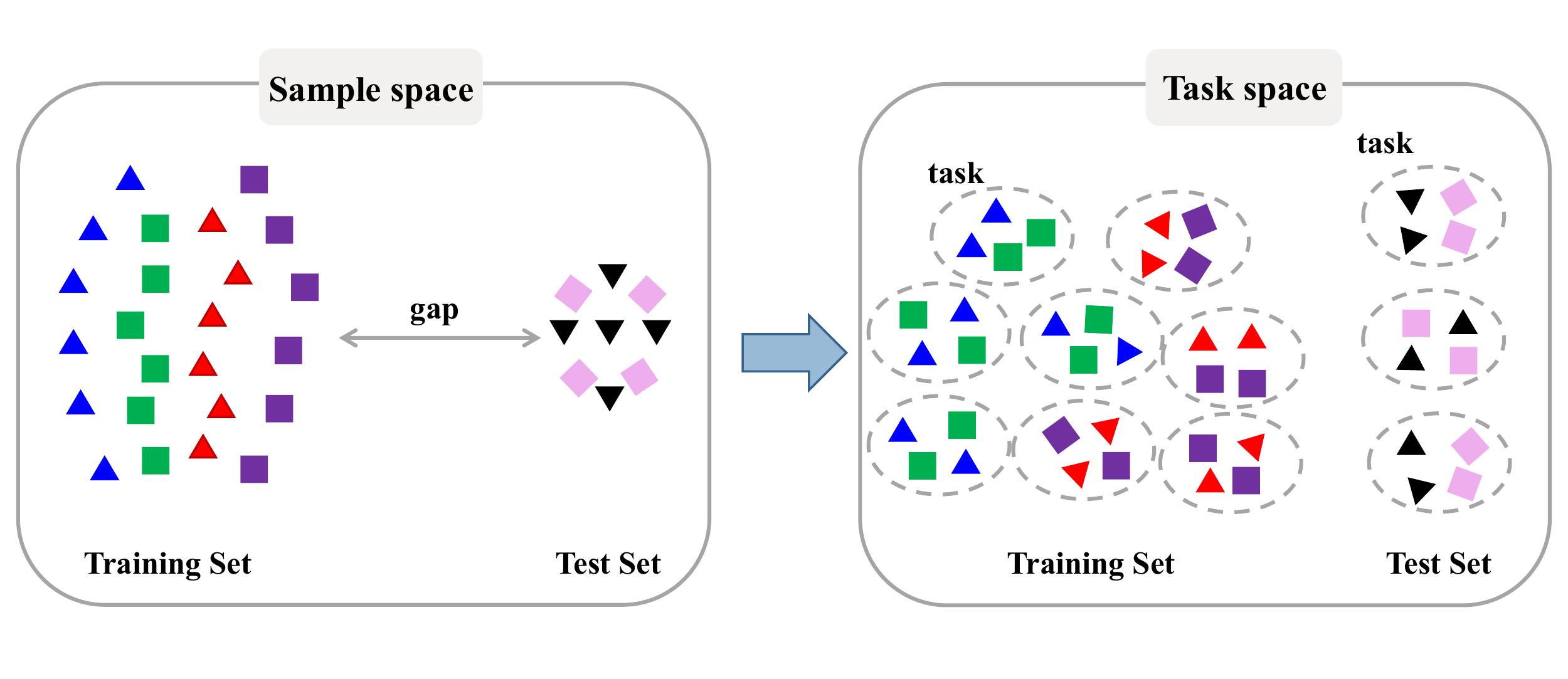}}
 \caption{Converting the distribution consistency assumption in the sample level into the task level. Each geometric shape indicates one sample and each color represents one class.}
 \label{fig:DistributionIllus}
 \end{center}
 \vskip -0.3in
 \end{figure}

 Generic zero-shot learning models usually make a distribution consistency assumption between the training and test sets (\emph{i.e.}, independent and identically distributed assumption), thus guaranteeing the model trained on the training set can generalize to the test set. However, because of the existence of unseen classes, the sample distribution of training set is relatively different from that of test set. Then, the generalization performance on the test set cannot be well guaranteed. As an illustration, from the left side of Figure~\ref{fig:DistributionIllus}, we can observe that there may exist a large distribution gap between training and test sets in the sample space because these two sets are class-disjoint, making it weakly transferable. We cannot even estimate the gap due to a complete lack of training samples in the unseen class set $Y^u$.
 Fortunately, we can assume the above distribution consistency in a task space instead of the sample space (see the right side of Figure~\ref{fig:DistributionIllus}). The task space is composed of a lot of similar tasks, \emph{e.g.}, zero-shot tasks. From the perspective of task-level distribution, we can construct a large number of zero-shot tasks within the training set $\mathcal{D}^{tr}$, by simulating the target zero-shot task. Consequently, the sample distribution gap can be mitigated.
 
 %--------------------Figure 4---------------------------%
 %--------------------Figure 4---------------------------%

 According to the above task-level distribution consistency assumption, we propose the following episode-based convolutional prototype learning framework.
 Specifically, let $\{\left \langle \mathcal{X}_{1}^{tr},\bm A_{1}^{s}\right \rangle,\cdots,\left \langle \mathcal{X}_{n}^{tr},\bm A_{n}^{s}\right \rangle\}$
%  $\{\mathcal{D}_{1}^{tr},\cdots,\mathcal{D}_{n}^{tr}\}$
 be a set of zero-shot tasks randomly sampled from the training set $\mathcal{D}^{tr}$. Then our episode-based convolutional prototype learning framework (CPL) can be formulated as follows:
 
 \begin{align} \label{Objective}
    \begin{split}
      &  \Gamma =\arg \min _{\theta}\sum_{i=1}^{n}\sum _{ x \in \mathcal{X}_{i}^{tr}} \mathcal{L}( x, \bm A_{i}^{s}, y_{true}),
    \end{split}
\end{align}
 where $\theta$ denotes the parameter set of our convolutional prototype learning model. The core idea here is to simulate the target zero-shot task and conduct a lot of similar zero-shot tasks with training set. In this way, we can build a task-based space to mend the gap among different sample distributions. 
 It is worth noting that the classes covered by each task $\left \langle \mathcal{X}_{i}^{tr},\bm A_{i}^{s}\right \rangle$ should be as disjoint as possible. This can facilitate better transferable knowledge for new tasks.
 At each training step, we generate discriminative visual prototypes inside each sampled task according to the current model. 
 
  %--------------------Figure 5---------------------------%
 \begin{figure*}[t]
\begin{center}
   \includegraphics[width=6.0in ]{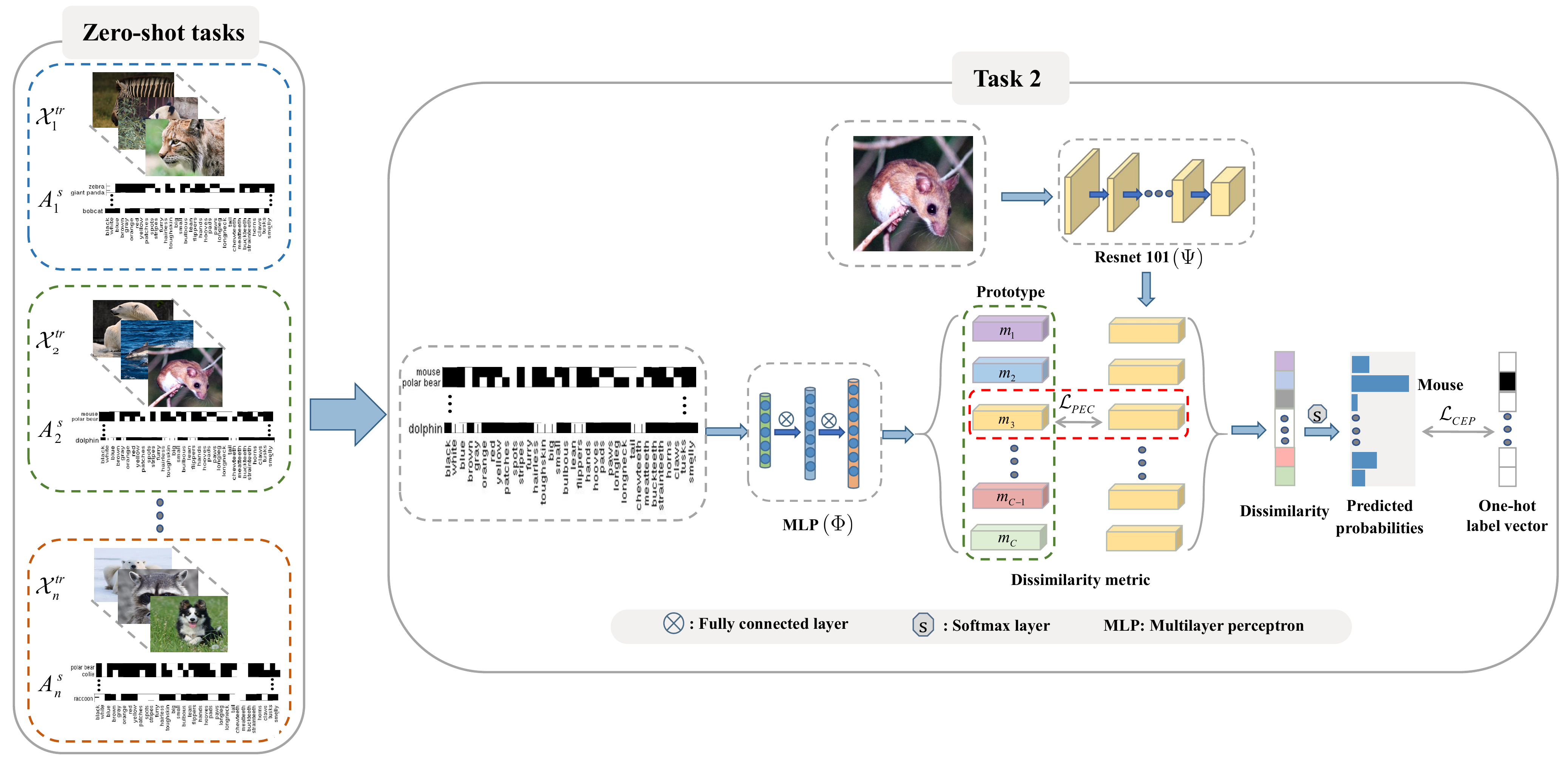}
\end{center} %\linewidth
   \caption{The illustration of our proposed CPL framework for zero-shot recognition.}
\label{fig:framework}
\end{figure*}
 %--------------------Figure 5---------------------------%
 
 More specifically, for discriminative recognition, the minimization of $\mathcal{L}(\cdot)$ is encouraged to achieve \underline{two goals} of i) minimizing the classification error of $x$ via visual prototypes (CEP); ii) minimizing the encoding cost of $x$ via visual prototypes (PEC).
 For this end, as shown in Figure~\ref{fig:framework}, we consider a decomposition of the objective function in Eq.~(\ref{Objective}) into two functions, corresponding to the two aforementioned objectives, as
 \begin{align} \label{Obj_decomp}
    \begin{split}
      &  \mathcal{L}( x, \bm A_{i}^{s}, y_{true})  \triangleq \\ & \lambda \mathcal{L}_{CEP}( x, \bm A_{i}^{s}, y_{true}) +  \mathcal{L}_{PEC}( x, \bm A_{i}^{s}, y_{true}),
    \end{split}
\end{align}
where $\lambda \in[ 0,1]$ balances the effects of two functions.

 The detailed definition about $\mathcal{L}_{CEP}(\cdot)$ and $\mathcal{L}_{PEC}(\cdot)$ is presented in Section~\ref{CEP} and Section~\ref{PEC}, respectively.
 
 %----------------------------------[loss 1]-----------------------------%
 \subsubsection{Classification error via prototypes (CEP)} \label{CEP}
 Suppose there exist $C$ classes and $S$ support samples per-class in the $i$-th sampled task $\left \langle \mathcal{X}_{i}^{tr},\bm A_{i}^{s}\right \rangle$. Then the sets of class attributes and classes are denoted as $\bm A_{i}^{s}=\{\bm a_{i_1}^{s}, \cdots, \bm a_{i_C}^{s}\}$ and $Y_{i}^{s}=\{s_{i_1}^{s}, \cdots, s_{i_C}^{s}\}$, respectively.
 In our CPL model, we need to learn $C$ visual prototypes, denoted as $\bm M = \{\bm m_{1},\cdots,\bm m_{C}\}$, to represent such $C$ classes in visual space. 
 Once we obtain $\bm M$, the probability of sample $x$ in $\mathcal{X}_{i}^{tr}$ belonging to the $j$-th prototype $\bm m_j$ has the following relationship with the similarity between $x$ and $\bm m_j$:
  \begin{align} \label{probability}
    \begin{split}
      &  p(x \in \bm m_j|x) \propto - ||\Psi(x)-\bm m_j||_2,
    \end{split}
  \end{align}
where $\bm m_j = \Phi(\bm a_{i_j}^{s})$. That is, we learn visual prototypes from class attributes via a non-linear function $\Phi$ as illustrated in Figure~\ref{fig:framework}. Actually, this is inspired by multi-view learning~\cite{xu2013survey}.
Then, the above probability can be further normalized as
  \begin{align} \label{ProbabilityNormalize}
    \begin{split}
     & p(x \in \bm m_j|x) = \frac{\exp^{-\gamma ||\Psi(x)-\bm m_j||_2}}{\sum^{C}_{l=1}\exp^{-\gamma  ||\Psi(x)-\bm m_l||_2}},
    \end{split}
 \end{align}
where $\gamma$ is the temperature to mitigate overfitting~\cite{hinton2015distilling}, and $\gamma=1$ is a common option. In particular, $\gamma$ “softens” the softmax (raises the output entropy) with $\gamma<1$. As $\gamma \rightarrow 0$, %$p\left(c_{j}^{\textrm{s}}|x_{i}^{\textrm{tr}}\right)\rightarrow
$p(x \in \bm m_j|x)\rightarrow
1/C$, which leads to maximum uncertainty. As $\gamma \rightarrow \infty$, the probability collapses to a point mass (\emph{i.e.}, 
$p(x \in \bm m_j|x)=1$). 
Since $\gamma$ does not change the maximum of the softmax function, the class prediction remains unchanged if $\gamma \neq 1$ is applied after convergence.
Plugging the probability in Eq.~(\ref{ProbabilityNormalize}) into cross-entropy loss over the training sample $x$ yields
 \begin{align} \label{CEPloss}
    \begin{split}
     & \mathcal{L}_{CEP}( x, \bm A_{i}^{s}, y_{true})=-\sum \limits_{j=1}^{C} q_{j} \log p(x \in \bm m_j|x),
    \end{split}
\end{align}
where $q_{j}=1$ if $y_{true}=s^s_{i_{j}}$ and 0 otherwise.

Consequently, we can guarantee the classification accuracy of all the samples effectively by learning discriminative prototypes.

%----------------------------------[losss 2]-----------------------------%
\subsubsection{Prototype encoding cost (PEC)}  \label{PEC}
Furthermore, to make prototype learning generalize more smoothly on new tasks, we should improve the representativeness of learned prototypes in visual space.
% In essence, each sample can find its prototype. 
% By minimizing the encoding cost of samples, we
For this end, we additionally minimizing the prototype encoding cost as follows
 \begin{align} \label{PECloss}
    \begin{split}
     & \mathcal{L}_{PEC}( x, \bm A_{i}^{s}, y_{true})=||\Psi(x)-
     \bm m_{y_{true}}||_2.
    \end{split}
 \end{align}
where $\bm m_{y_{true}}$ is the prototype of the $y_{true}$-th class, corresponding to the label of sample $x$.

\subsection{CPL: recognition}\label{ModelPrediction}
During the test phase, we have such a zero-shot task $\left \langle \mathcal{X}^{te}, \bm A^{u}\right \rangle$ with the unseen class set $Y^{u}$. To recognize the label of any one sample $x^{te}$ in $\mathcal{X}^{te}$, we first need to obtain visual prototypes of all the classes covered in $Y^{u}$ via the function $\Phi$ learned during the training phase. And then, we compare the sample $x^{te}$ with all prototypes and classify it to the nearest prototype.
As a result, the label of sample $x^{te}$, denoted $y^{te}$, is predicted as follows
\begin{align} \label{Recog_Pro}
    \begin{split}
     & y^{te} = \arg \max_{u_{k}\in Y^{u}} p(x^{te} \in \bm m_k^{te}|x^{te}),
    \end{split}
\end{align}
where $\bm m_k^{te} = \Phi(\bm a_{k}^{u})$.

\section{Experimental results and analysis}
In this section, we first detail our experimental protocol, and then present the experimental results by comparing our CPL with the state of the art methods for zero-shot recognition on four benchmark datasets under various settings.

 %--------------------TABLE 1---------------------------%
\begin{table*}[t]
% \vskip 0.15in
\begin{center}
    \begin{tabular}{|cc|cccc|ccc|}
    \hline
    \multicolumn{2}{|c|}{} & \multicolumn{4}{c|}{\textbf{Number of Classes}}  & \multicolumn{3}{c}{\textbf{Number of Images}}\\
     \textbf{Dataset} & \# \textbf{attributes} & \textbf{Total} & \# \textbf{training}  & \# \textbf{validation} & \# \textbf{test} & \textbf{Total} & \# \textbf{training} & \# \textbf{test}  \\
     \hline
     SUN   & 102 & 717 & 580 & 65 & 72 & 14340  & 10320 & 2580+1440\\
     AWA2  & 85  & 50  & 27  & 13 & 10 & 37322  & 23527 & 5882+7913 \\
     CUB   & 312 & 200 & 100 & 50 & 50 & 11788  & 7057  & 1764+2967 \\
     aPY   & 64  & 32  & 15  & 5  & 12 & 15339  & 5932  & 1483+7924\\
    \hline
    \end{tabular}
\end{center}
\caption{Statistics for four datasets. Note that test images include the images in both seen and unseen class domains.} 
\label{tab:datasets}
% \vskip -0.2in
\end{table*}
 %--------------------TABLE 1---------------------------%
 
\subsection{Evaluation setup and metrics}
\paragraph{Datasets.} Among the most widely used datasets for ZSL, we select four attribute datasets. Two of them are coarse-grained, one small (aPascal \& Yahoo (aPY)~\cite{farhadi2009describing}) and one medium-scale (Animals with Attributes (AWA2)~\cite{xian2018zero}). Another two datasets (SUN Attribute (SUN)~\cite{patterson2012sun} and CUB-200-2011 Birds (CUB)~\cite{wah2011caltech}) are both fine-grained and medium-scale. Details of all dataset statistics are in Table~\ref{tab:datasets}.

\paragraph{Protocols.} 
We adopt the novel rigorous protocol\footnote{\url{http://www.mpi-inf.mpg.de/zsl-benchmark}} proposed in~\cite{xian2018zero}, insuring that none of the unseen classes appear in ILSVRC 2012 1K, since ILSVRC 2012 1K is used to pre-train the Resnet model. Otherwise the zero-shot rule would be violated.
In particular, this protocol involves two settings: \emph{Standard ZSL} and \emph{Generalized ZSL}. 
The latter emerges recently under which the test set contains data samples from both seen and unseen classes. This setting is thus clearly more reflective of real-world application scenarios. By contrast, the test set in standard ZSL only contains data samples from the unseen classes.

% \paragraph{Episode composition}
% To build different meta-tasks, we randomly choose $N_C$ classes and $N_S$ samples per-class for each episode training, and $N_C$ should less than the number of seen class. On one hand, the search space is inconsistent for different meta-tasks, so the learned model can be well generalized to unseen class, as new classes always appear in each episode training. On the other hand, we aim to learn the meta-knowledge about meta-tasks which conform the same distribution, and then our method can be well transfer to the target task. In addition, To satisfy the target task, we set $N_C$ to the number of unseen class. In our experiments, we showed that this setup is the most effective and performs best.

\paragraph{Evaluation metrics.} 
During test phase, we are interested in having high performance on both densely and sparsely populated classes. Thus, we use the unified evaluation protocol proposed in~\cite{xian2018zero}, where the average accuracy is computed independently for each class. Specifically, under the standard ZSL setting, we measure average per-class top-1 accuracy by
\begin{align*}
Acc_{\mathcal{U}}=\frac{1}{L}\sum_{i=1}^{L}\frac{\text{\# correct predictions in}~u_{i}}{\text{\# samples in}~u_{i}}.
\end{align*}

While under the generalized ZSL setting, we compute the harmonic mean ($H$) of $Acc_{\mathcal{S}}$ and $Acc_{\mathcal{U}}$ to favor high accuracy on both seen and unseen classes:
\begin{align*}
H=\frac{2*Acc_{\mathcal{S}}*Acc_{\mathcal{U}}}{Acc_{\mathcal{S}}+Acc_{\mathcal{U}}},
\end{align*}
where $Acc_{\mathcal{S}}$ and $Acc_{\mathcal{U}}$ are the accuracy of recognizing the test samples from the seen and unseen classes respectively, and 
\begin{align*}
Acc_{\mathcal{S}}=\frac{1}{K}\sum_{i=1}^{K}\frac{\text{\# correct predictions in}~s_{i}}{\text{\# samples in}~s_{i}}.
\end{align*}

\paragraph{Implementation details.}
By tuning on validation set, we set $C$ to be the same with the number of unseen classes, and $S=10$. The parameter $\lambda$ is selected from $\left\{0.05,0.1,0.2,1\right\}$ and $\gamma$ is selected from $\left\{0.9,0.95\right\}$. For each dataset, our CPL is trained for 40 epochs with weight decay $10^{-5}$ on two fine-grained datasets and $10^{-4}$ on two coarse-grained datasets. Meanwhile, the learning rate 
is initialized with Adam by selecting from $\left\{0.00003,0.00005,0.0002\right\}$.   Specially, i) we adopt Resnet101 as embedding module that is pre-trained on ILSVRC 2012 1k without fine-tuning. The input to Resnet101 is a color image that is first normalized with means $\left\{0.485,0.456,0.406\right\}$ and standard deviations $\left\{0.229,0.224,0.225\right\}$ with respect to each channel. ii) We utilize a MLP network as attribute embedding module. The size of hidden layer (as in Fig.~\ref{fig:framework}) is set to 1200 for CUB and 1024 for another three datasets, and the output layer is set to the same size (2048) as image embedding module for all the datasets. In addition, we add weight decay ($\ell_{2}$ regularisation), and ReLU non-linearity for both hidden and output layers.

\paragraph{Compared methods.} We choose to compare with a wide range of competitive and representative inductive ZSL approaches, especially those that have achieved the state-of-the-art results recently. In particular, such compared approaches involve both shallow and deep models.

 %--------------------TABLE 2---------------------------%
\begin{table}[htb!]
\begin{center}
\begin{tabular}{|c|c|c|c|c|}
\hline
    {\bfseries Method} & {\bfseries SUN } & {\bfseries AWA2} & {\bfseries CUB} & {\bfseries aPY} \\
    \hline

      DAP \cite{lampert2013attribute}        & 39.9  & 46.1  & 40.0  & 33.8 \\
      IAP \cite{lampert2013attribute}        & 19.4  & 35.9  & 24.0  & 36.6 \\
      CONSE \cite{norouzi2013zero}           & 38.8  & 44.5  & 34.3  & 26.9 \\
      CMT \cite{socher2013zero}              & 39.9  & 37.9  & 34.6  & 28.0 \\
      SSE \cite{zhang2015zero}               & 51.5  & 61.0  & 43.9  & 34.0 \\
      LATEM \cite{Xian_2016_CVPR}            & 55.3  & 55.8  & 49.3  & 35.2 \\
      DEVISE \cite{frome2013devise}          & 56.5  & 59.7  & 52.0  & 39.8 \\
      SJE \cite{akata2015evaluation}         & 53.7  & 61.9  & 53.9  & 32.9 \\
      SYNC \cite{changpinyo2016synthesized}  & 56.3  & 46.6  & 55.6  & 23.9 \\
      SAE \cite{kodirov2017semantic}         & 40.3  & 54.1  & 33.3  & 8.3  \\
      DEM \cite{zhang2017learning}           & 61.9  & 67.1  & 51.7  & 35.0 \\
      PSR \cite{annadani2018preserving}      & 61.4  & 63.8  & 56.0  & 38.4 \\
      DLFZRL \cite{Tong_2019_CVPR}           & 59.3  & 63.7  & {\bfseries 57.8}  & 44.5 \\
    \hline
      {\bfseries CPL} & {\bfseries 62.2}  & {\bfseries 72.7} & 56.4 & {\bfseries 45.3}\\ 
   \hline
\end{tabular}
\end{center}
\caption{Comparative results of standard zero-shot learning on four datasets.}
\label{tab:standard_results}
\end{table}
 %--------------------TABLE 2---------------------------%

%--------------------TABLE 3---------------------------%
\begin{table*}[htb!]
\begin{center}
\begin{tabular}{|c|c|c|c|c|c|c|c|c|c|c|c|c|}
\hline
\multicolumn{1}{|c|}{ } & \multicolumn{3}{|c|}{\bfseries SUN} & \multicolumn{3}{|c|}{\bfseries AWA2} & \multicolumn{3}{|c|}{\bfseries CUB} & \multicolumn{3}{|c|}{\bfseries aPY}\\
\hline
{\bfseries Method} &$Acc_{\mathcal{U}}$ & $Acc_{\mathcal{S}}$ & H  & $Acc_{\mathcal{U}}$ & $Acc_{\mathcal{S}}$ & H  & $Acc_{\mathcal{U}}$ & $Acc_{\mathcal{S}}$ & H & $Acc_{\mathcal{U}}$ & $Acc_{\mathcal{S}}$ & H\\
\hline
DAP \cite{lampert2013attribute}       & 4.2   & 25.1  & 7.2   & 0.0   & 84.7  & 0.0   & 1.7   & 67.9  & 3.3   & 4.8   & 78.3  & 9.0 \\
IAP \cite{lampert2013attribute}       & 1.0   & 37.8  & 1.8   & 0.9   & 87.6  & 1.8   & 0.2   & {\bfseries 72.8}  & 0.4   & 5.7   & 65.6  & 10.4 \\
CONSE \cite{norouzi2013zero}          & 6.8   & 39.9  & 11.6  & 0.5  & {\bfseries 90.6}  & 1.0   & 1.6   & 72.2  & 3.1   & 0.0   & {\bfseries91.2}  & 0.0 \\
CMT \cite{socher2013zero}             & 8.1   & 21.8  & 11.8  & 0.5   & 90.0  & 1.0   & 7.2   & 49.8  & 12.6  & 1.4   & 85.2  & 2.8 \\
SSE \cite{zhang2015zero}              & 2.1   & 36.4  & 4.0   & 8.1   & 82.5  & 14.8  & 8.5   & 46.9  & 14.4  & 0.2   & 78.9  & 0.4 \\
LATEM \cite{Xian_2016_CVPR}           & 14.7  & 28.8  & 19.5  & 11.5  & 77.3  & 20.0  & 15.2  & 57.3  & 24.0  & 0.1   & 73.0  & 0.2 \\
DEVISE \cite{frome2013devise}         & 16.9  & 27.4  & 20.9  & 17.1  & 74.7  & 27.8  & 23.8  & 53.0  & 32.8  & 4.9   & 76.9  & 9.2 \\
SJE \cite{akata2015evaluation}        & 14.7  & 30.5  & 19.8  & 8.0   & 73.9  & 14.4  & 23.5  & 59.2  & 33.6  & 3.7   & 55.7  & 6.9 \\
SYNC \cite{changpinyo2016synthesized} & 7.9   & {\bfseries 43.3}  & 13.4  & 10.0  & 90.5  & 18.0  & 11.5  & 70.9  & 19.8  & 7.4   & 66.3  & 13.3 \\
SAE \cite{kodirov2017semantic}        & 8.8   & 18.0  & 11.8  & 1.1   & 82.2  & 2.2   & 7.8   & 54.0  & 13.6  & 0.4   & 80.9  & 0.9  \\
DEM \cite{zhang2017learning}          & 20.5  & 34.3  & 25.6  & 30.5  & 86.4  & 45.1  & 19.6  & 57.9  & 29.2  & 11.1  & 75.1  & 19.4 \\
PSR \cite{annadani2018preserving}     & 20.8  & 37.2  & {\bfseries 26.7}  & 20.7  & 73.8  & 32.3  & 24.6  & 54.3  & 33.9  & 13.5  & 51.4  & 21.4 \\
DLFZRL \cite{Tong_2019_CVPR}          & -  & - & 24.6 & - & - & 45.1 & - & - & 37.1 & - & - & {\bfseries 31.0} \\
\hline
{\bfseries CPL}                 & {\bfseries 21.9}  & 32.4  & 26.1  & {\bfseries 51.0}  & 83.1  & {\bfseries 63.2}  & {\bfseries 28.0}  & 58.6  & {\bfseries 37.9}  & {\bfseries 19.6}  & 73.2  & 30.9 \\
\hline
\end{tabular}
\end{center}
\caption{Comparative results of generalized zero-shot learning on four datasets.}
\label{tab:generalized_results}
\end{table*}
 %--------------------TABLE 3---------------------------%

{\subsection{Standard ZSL}}
We firstly compare our CPL method with several state-of-the-art ZSL approaches under the standard setting. The comparative results on four datasets are shown in Table~\ref{tab:standard_results}. It can be observed that: i) Our model consistently performs best on all the datasets except CUB, validating that our episode-based convolutional prototype learning framework is indeed effective for zero-shot recognition tasks.
ii) For the three datasets (e.g., {SUN}, {AWA2}, and {aPY}), the improvements obtained by our model over the strongest competitor range from 0.3\% to 5.6\%. This actually creates new baselines in the area of ZSL, given that most of the compared approaches take far more complicated image generation networks and some of them even combine two or more feature/semantic spaces. 
iii) In particular, for the two coarse-grained datasets ({aPY} and {AWA2}), our CPL achieves 0.5\% and 5.6\% significant improvements over the strongest competitors, showing its great advantage in coarse-grained object recognition problems. 
iv) Although DLFZRL~\cite{Tong_2019_CVPR} performs better than our CPL with 1.4\% on CUB, it still can be observed that our CPL takes more advantage on most cases.

We further discuss the reasons why our CPL outperforms existing methods on zero-shot recognition tasks. First, compared with DEM~\cite{zhang2017learning} that conducts a non-linear neural network to model the relationship between the visual and the semantic space, our model also possesses such a non-linear property.
Besides, our framework improves the discriminability and representativeness of learned prototypes by minimizing the classification error of seen class samples, and encoding cost of prototypes, at task-level respectively. Thus, recognition performances can be improved dramatically. 
Second, to achieve better performances on novel classes, PSR~\cite{annadani2018preserving} proposes to preserve semantic relations in the semantic space when learning embedding function from the semantic to the visual space. However, it is hard to select an appropriate threshold to define the semantically similar and dissimilar classes in our daily life. By contrast, our framework needs not to predefine these relations. 
% Our CPL improves the representativeness of learned prototypes in visual space by minimizing the prototype encoding cost, which can guarantee prototype learning generalize more smoothly on new tasks.
% Especially,  it can be seen that our model can better generalize to novel classes than the compared methods, this verifies that our CPL can capable of transferring knowledge smoothly to recognize unseen samples based on the distribution consistency assumption at task-level. 
Third, a recent work DLFZRL~\cite{Tong_2019_CVPR} aims to learn discriminative and generalizable representations from image features and then improves the performance of existing methods. In particular, they utilized DEVISE~\cite{frome2013devise} as the embedding function and obtained very competitive results on four benchmark datasets. Unlike DLFZRL~\cite{Tong_2019_CVPR}, our CPL has different motivations, and then provides the learned representations with a novel concept (i.e., visual prototypes).
% Although our CPL has different motivations from this approach, we provide new application framework for the learned representations, which is very helpful in solving the zero-shot recognition tasks.

% The comparative results on four benchmarks under standard ZSL setting are shown in Table \ref{t2}. For comprehensive comparison with the inductive state-of-the-art ZSL models, It can be seen that: 1) Our model performs the best on four datasets, validating the reasonableness of our key motivation for the design of CPL. 2) As shown in the Table \ref{t2}, our approach achieves a mean class accuracy of 62.2\% on SUN which is best result among the comparison methods. In addition, we obtain 72.7\% on AWA2 dataset, better than the next best method by nearly 5.6\%. On CUB, proposed CPL achieves a best accuracy of 56.4\%. On aPY dataset, we obtains a 5.8\% improvement over the state-of-the-art PSR~\cite{annadani2018preserving}. This actually creates new insights in the area of ZSL, given that most of the compared models based on the distribution consistency assumption between seen and unseen domains at sample-level.  

\subsection{Generalized ZSL}

In real applications, whether a sample is from a seen or unseen class is unknown in advance. Hence, generalized ZSL is a more practical and challenging task compared with standard ZSL. Here, we further evaluate the proposed model under the generalized ZSL setting.
The compared approaches are consistent with those in standard ZSL.
Table~\ref{tab:generalized_results} reports the comparative results, much lower than those in standard ZSL. This is not surprising since the seen classes are included in the search space which act as distractors for the samples that come from unseen classes. 
Additionally, it can be observed that generally our method improves the overall performance (i.e., harmonic mean $H$) over the strongest competitor by an obvious margin (0.8\% $\sim$ 18.1\%). 
Such promising performance boost mainly comes from the improvement of mean class accuracy on the unseen classes, meanwhile without much performance degradation on the seen classes. Especially, our method obtains the best performance for recognizing unseen samples on four benchmarks, the improvements range from 1.1\% to 20.5\%. These compelling results also verify that our method can significantly alleviate the strong bias towards seen classes. 
This is mainly due to the fact that: i) Our CPL significantly improves the representability and discriminability of learned prototypes by means of the corresponding objective functions.
ii) Unlike most of existing inductive approaches (e.g., DEM~\cite{zhang2017learning}) that assumes distribution consistency between seen and unseen classes at sample-level, our CPL creatively makes an assumption at task-level). 
Consequently, our CPL can learn more transferable knowledge that is able to generalize well to the unseen class domain.

% Compared with standard ZSL, it is much more technically difficult to address generalized ZSL due to the search space contains both seen and unseen classes. The comparative results on four benchmarks are presented in Table \ref{t3}, where our model is compared with the existing approaches. We have the following observations: 1) Different ZSL methods have a different trade-off between the seen and unseen class accuracies, and the overall performance is thus measured by harmonic mean. 2) With respect to the state-of-the-art, our approach achieves a harmonic mean accuracy of 26.1\% on SUN which is the second best among all the reported approaches. In addition, we achieve 63.2\% on the AWA2 dataset, better than the next best method by nearly 18.1\%. On CUB, our method achieves a best harmonic mean accuracy of 37.9\%. On aPY dataset, our methods obtains a 6.3\% improvement comparison with the next best method. This verifies that our model have the strong generalization ability under generalized ZSL setting. 

{\subsection{Ablation study}}

\begin{figure}[t]
\begin{center}
   \includegraphics[width=0.8\linewidth]{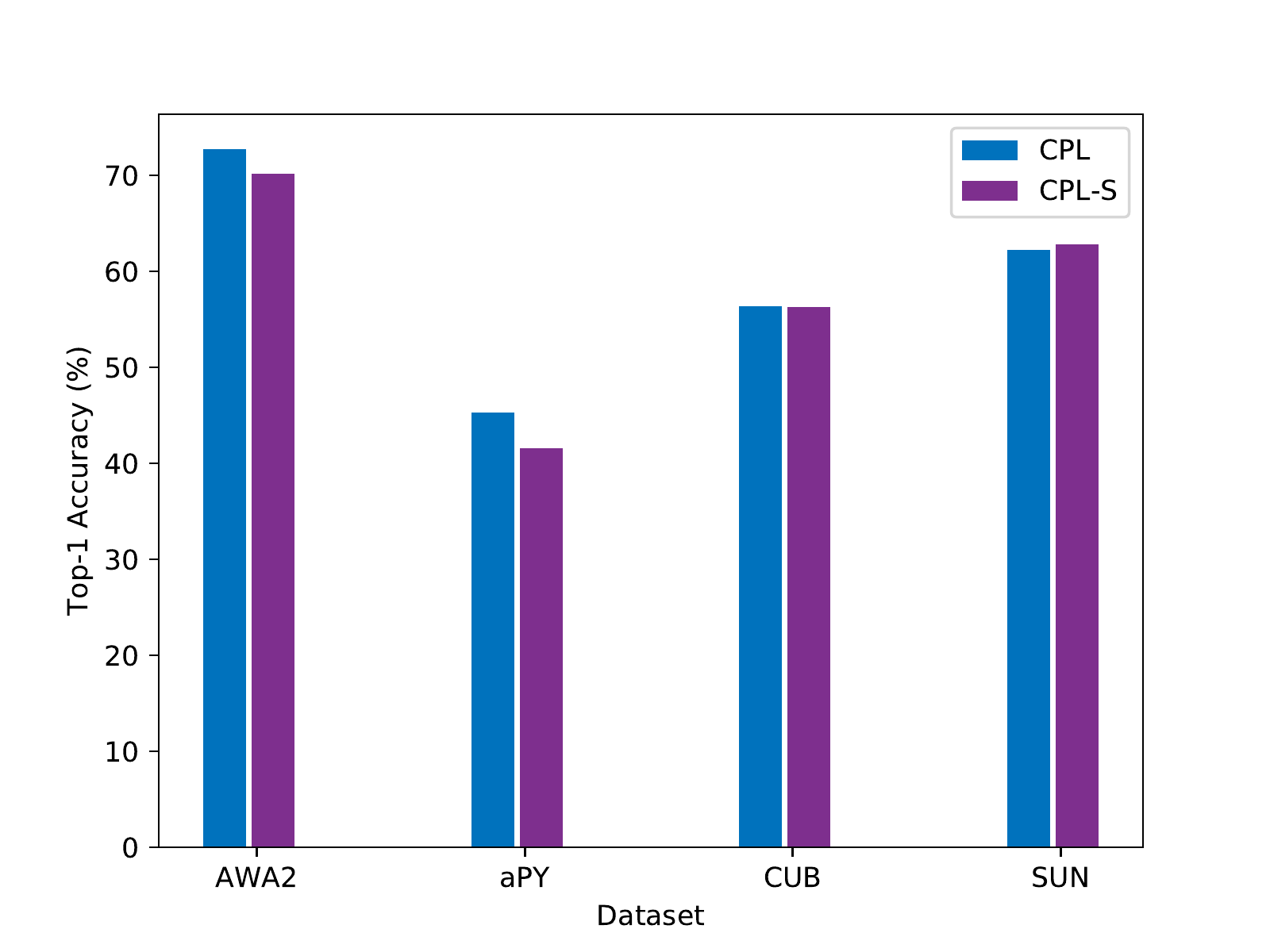}
\end{center}
   \caption{The comparison results of task-level and sample-level.}
\label{fig:CPL_TS}
\end{figure}

% \paragraph{Effectiveness of Episode Training}  %% episode training or meta learning
\paragraph{Effectiveness of distribution consistency at task-level.}  %% episode training or meta learning
First, we evaluate the assumption of distribution consistency at task-level.
In this experiment, CPL denotes the assumption of distribution consistency at task-level which randomly choices $C$ classes and $S$ support samples per-class for each zero-shot task, and here we set $C$ to be the same with the number of unseen classes and $S=10$. Let \textbf{CPL-S} denote the assumption of distribution consistency at sample-level that randomly selects a batch with the sample number $C \times S$ for each iteration. 
Then, we conduct corresponding experiments on four datasets and report the comparison results in Figure \ref{fig:CPL_TS}.
It can be concluded that: i) Our CPL performs better than CPL-S especially on two coarse-grained datasets. Concretely, on AWA2 dataset, CPL obtains a 2.5\% improvement compared with CPL-S. Meanwhile, our method also achieves a 2.6\% improvement on aPY dataset. Additionally, CPL and CPL-S have the basically same performance on two fine-grained datasets. That is to say, our CPL based on the assumption at task-level can generalize well to novel classes, and effectively mitigate the significant distribution gap at sample-level. ii) Significant differences between classes are more beneficial for CPL to learn transferable knowledge.

%--------------------Figure 6---------------------------%
 %--------------------Figure 6---------------------------%

 %--------------------TABLE 4---------------------------%
\begin{table}[tb!]
\begin{center}
\begin{tabular}{|c|c|c|c|c|c|c|}
\hline
{Value of \bfseries $C$} & 3 & 6 & 9 & 12 & 15 &  20\\
\hline
{\bfseries CPL} & 42.3 & 42.6 & 42.9 & {\bfseries 45.3} & 42.1 & 42.7 \\
\hline
\end{tabular}
\end{center}
\caption{Parameter influence of $C$ on ZSL task using aPY.}
\label{tab:C_influence}
\end{table}
 %--------------------TABLE 4---------------------------%

\paragraph{Effectiveness of $C$.}

Second, we discuss the influence of $C$ in our CPL framework. 
Under standard ZSL setting, we set different $ C = \left\{3, 6, 9, 12, 15, 20\right\} $ on aPY dataset, and $S=10$ is same for each experiment. This means that we build  the corresponding sets of zero-shot tasks for different experiments and each zero-shot task consists of $C$ seen classes. 
We report the results on Table \ref{tab:C_influence}, it can be seen that $C=12$ that is the same with the number of unseen classes outperforms than other setup with a 2.6\% improvement. Such comparative results further verify that our assumption of distribution consistency at task-level is more suitable for ZSL problem. In addition, the learned knowledge can be well transferred from the seen to the unseen classes. 

%--------------------TABLE 5---------------------------%
\begin{table}[tb!]
\begin{center}
\begin{tabular}{|c|c|c|c|c|c|}
\hline
{\bfseries Method} & {\bfseries SUN} & {\bfseries AWA2} & {\bfseries CUB} & {\bfseries aPY} \\
\hline
$B_1$ & 59.7 & 69.9 & 53.6 & 41.8 \\
\hline
$B_2$ & 60.1 & 70.0 & 48.3 & 39.2 \\
\hline
{\bfseries CPL} & {\bfseries 62.2} & {\bfseries 72.7} & {\bfseries 56.4} & {\bfseries 45.3} \\
\hline
\end{tabular}
\end{center}
\caption{Effectiveness of CEP and PEC functions.}
\label{tab:Effective_losses}
\end{table}
 %--------------------TABLE 5---------------------------%

\paragraph{Effectiveness of CEP and PEC functions.}

Finally, We perform the ablation experiments of loss functions under standard ZSL setting. We define two baseline settings to prove the importance of each objective function. 
For baseline {\bfseries $B_1$}, we set $\lambda = 0$ to evaluate the contribution of the distance based cross entropy loss. This results in the improvement of representativeness of learned prototypes in visual space. 
For baseline {\bfseries $B_2$}, only $CEP$ loss is used to learn the embedding model which helps us to better understand the importance of objective $PEC$. This effectively improves the classification accuracy of all the samples by learning discriminative prototypes.
The comparative results is shown in Table \ref{tab:Effective_losses}, it can be seen that: i)   The improvements obtained by introducing the $CEP$ loss range from 2.5\% to 3.5\%. This indicates that the objective $CEP$, which aims to improve the discriminability of learned prototypes, is beneficial for solving zero-shot recognition tasks. ii) The improvements range from 2.1\% to 8.1\% when including $PEC$ loss. This proves that improving the representativeness of learned prototypes can make our model better generalize to novel classes. iii) By improving both the discriminability and representativeness of learned prototypes, we achieve remarkable performance on zero-shot recognition tasks. 

\section{Conclusions and future work}
 In this paper, we propose a convolutional prototype learning (CPL) framework that is able to perform zero-shot recognition in visual space, and thus, avoid a series of problems caused by the provided class attributes.
 Meanwhile, the generalization ability of our CPL is significantly improved, by assuming distribution consistency between seen and unseen domains at task-level, instead of the popularly used sample-level.
 We have carried out extensive experiments about ZSL on four benchmarks, and the results demonstrate the obvious superiority of the proposed CPL to the state-of-the-art ZSL approaches. 
 It is also worth noting that the number of prototypes is fixed in our CPL. In essence, learning one prototype for a class is generally insufficient to recognize one class and differentiate two classes. Thus, our ongoing research work includes learning prototypes adaptively with the data distribution.

%-------------------------------------------------------------------------

{\small
\bibliographystyle{cvm}
\bibliography{cvmbib}
}

\end{document}